\documentclass[dvipsnames,format=sigconf]{acmart}

\usepackage{float}
\usepackage[ruled, linesnumbered ]{algorithm2e}
\usepackage{algpseudocode}
\usepackage{amsmath}

\usepackage{graphicx}
\usepackage{xcolor}
\usepackage{bm}
\usepackage{enumitem}

\AtBeginDocument{%
  \providecommand\BibTeX{{%
    \normalfont B\kern-0.5em{\scshape i\kern-0.25em b}\kern-0.8em\TeX}}}

\copyrightyear{2024} 
\acmYear{2024} 
\setcopyright{rightsretained} 
\acmConference[GECCO '24]{Genetic and Evolutionary Computation Conference}{July 14--18, 2024}{Melbourne, VIC, Australia}
\acmBooktitle{Genetic and Evolutionary Computation Conference (GECCO '24), July 14--18, 2024, Melbourne, VIC, Australia}
\acmDOI{10.1145/3638529.3654043}
\acmISBN{979-8-4007-0494-9/24/07}
\begin{document}


\title{Machine Learning-Enhanced Ant Colony Optimization for Column Generation}


\author{Hongjie Xu}
\affiliation{
  \institution{School of Computing Technologies, RMIT University}
  \city{Melbourne}
  \country{Australia}}
\email{s3880497@student.rmit.edu.au}
\author{Yunzhuang Shen}
\affiliation{
  \institution{University of Technology Sydney}
  \city{Sydney}
  \country{Australia}}
\email{yunzhuang.shen@uts.edu.au}
\author{Yuan Sun}
\affiliation{
  \institution{La Trobe Business School, La Trobe University}
  \city{Melbourne}
  \country{Australia}}
\email{yuan.sun@latrobe.edu.au}
\author{Xiaodong Li}
\affiliation{
  \institution{School of Computing Technologies, RMIT University}
  \city{Melbourne}
  \country{Australia}}
\email{xiaodong.li@rmit.edu.au}








\begin{abstract}

Column generation (CG) is a powerful technique for solving optimization problems that involve a large number of variables or columns. This technique begins by solving a smaller problem with a subset of columns and gradually generates additional columns as needed. However, the generation of columns often requires solving difficult subproblems repeatedly, which can be a bottleneck for CG. To address this challenge, we propose a novel method called machine learning enhanced ant colony optimization (MLACO), to efficiently generate multiple high-quality columns from a subproblem. Specifically, we train a ML model to predict the optimal solution of a subproblem, and then integrate this ML prediction into the probabilistic model of ACO to sample multiple high-quality columns. Our experimental results on the bin packing problem with conflicts show that the MLACO method significantly improves the performance of CG compared to several state-of-the-art methods. Furthermore, when our method is incorporated into a Branch-and-Price method, it leads to a significant reduction in solution time.

\end{abstract}

%
%




\maketitle

\setlength{\textfloatsep}{5pt}
\setlength{\floatsep}{2pt}

\section{Introduction}

Column generation (CG) is a powerful method for solving linear programs (LP) that have a large number of variables (or columns)~\citep{lubbecke2005selected}. It is commonly used to obtain tight LP bounds to accelerate the process of the branch-and-bound method in combinatorial optimization~\citep{barnhart1998branch}. CG is especially beneficial for tackling optimization problems that have a decomposable structure, such as vehicle routing, bin packing, and graph coloring problems.

CG solves a large-scale LP in iterative steps, starting from the LP containing a subset of columns, i.e., the restricted master problem (RMP). In an iteration, CG solves the current RMP and uses its dual solution to generate new columns that can improve the current RMP. Such columns should have negative reduced costs, and finding them typically involves solving an NP-hard subproblem called pricing problem. At optimality, no column with negative reduced costs can be further generated, and existing columns with nonzero values form an optimal solution to the original large-scale LP. 

Repeatedly solving pricing problems is often a bottleneck in CG~\citep{lubbecke2005selected}, and researchers have devised different approaches to tackle this issue, including exact methods, heuristics, and metaheuristics. It is widely recognized that the performance of CG is heavily influenced by both the quality and quantity of generated columns~\citep{enhanceYZ}. This differs from solving a traditional optimization problem, where typically only a single optimal column is sought.

\begin{figure}[!tb]
    \raggedleft
    \includegraphics[scale=0.45]{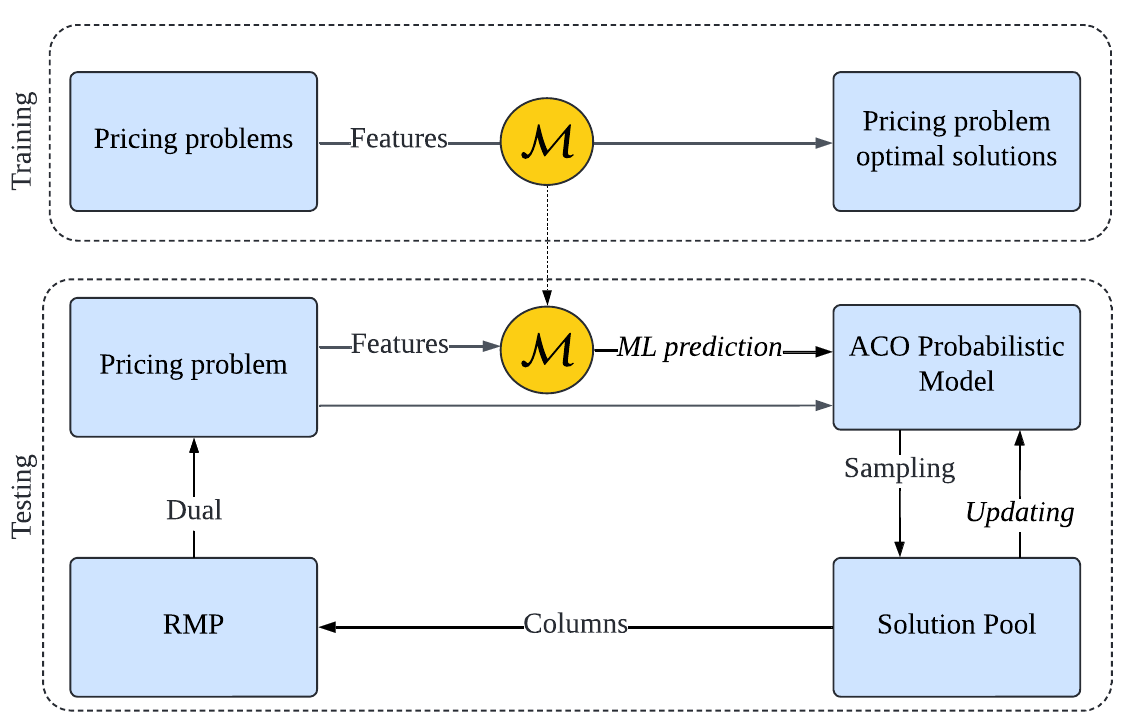}
    \caption{An illustration of our MLACO method as a pricing heuristic for CG. We train a ML model $\mathcal{M}$ on a set of solved pricing problem instances to learn a mapping from problem features to the optimal solutions. In the testing phase, we use the offline-trained ML model $\mathcal{M}$ to predict an optimal solution to an unseen pricing problem. The ML prediction is then incorporated into the ACO probabilistic model to sample multiple high-quality solutions, which are then iteratively improved in an online manner. 
    }
    \label{fig:mlaco}
\end{figure}

In this paper, we introduce a hybrid method called MLACO, which combines machine learning (ML) and ant colony optimization (ACO) to efficiently generate multiple high-quality columns, as illustrated in Figure~\ref{fig:mlaco}. In our method, we first train a ML model using a set of solved pricing problem instances. This ML model learns how to map from a set of problem-specific features and statistical measures to optimal solutions. Given a new problem instance, we use the ML model to predict its optimal solution and then incorporate the ML prediction into the ACO probabilistic model to generate a diverse set of high-quality columns. Our MLACO method is used to address the pricing problem and generate a diverse set of columns at every iteration of CG to accelerate its progress.

Our proposed MLACO method has the following advantages compared to existing methods for generating columns. Compared to exact and heuristic approaches based on mixed integer programming, our method can efficiently generate a large number of columns with a negative reduced cost. Compared to metaheuristics such as ACO, our method learns from optimally solved pricing problems and can generate high-quality columns more quickly. Compared to a pure ML-based sampling method~\citep{enhanceYZ}, our method can further improve the quality of the columns generated due to the online learning nature of ACO. 


Our main contributions can be summarized as follows.
\begin{itemize}[nolistsep]
    \item We propose the MLACO method for solving pricing problems in the context of CG. Our method uses ML to predict optimal solutions to a pricing problem and uses the ML prediction to accelerate ACO to efficiently generate multiple high-quality columns.
    \item We evaluate our method on the bin packing problem with conflicts (BPPC) and show that our method significantly accelerates CG compared to several baselines. Furthermore, we show that our MLACO method can tackle problems that are larger and more difficult than those used in training.
    \item We also evaluate our MLACO method within branch-and-price, an exact method that integrates CG to solve integer BPPC. We show that CG with our proposed method can reduce the solution time of Branch-and-Price.
\end{itemize}

\section{Background and Related Work}

\subsection{Problem Formulations}


We are given a set of bins $k \in \mathcal{K}$ with a uniform capacity $W$, and a set of items $i \in \mathcal{V}$, each with a non-negative weight $w_i$. Let $\mathcal{G}(\mathcal{V}, \mathcal{E})$ be a conflict graph, where a vertex represents an item and an edge indicates that the corresponding vertices have conflicts. The goal of the bin packing problem with conflicts (BPPC) is to place items in the minimum number of bins within the capacity limit, such that conflicting items are not placed in the same bin. This problem has various practical applications~\cite{renault2015online}, such as job scheduling, parallel computing, and database storage. 

Let the binary variable $x_{i,k}$ denote if item $i$ is assigned to bin $k$ and binary variable $y_{k}$ represent whether a bin $k$ is used. The BPPC can be formulated as an integer programming (IP) problem: 
\begin{align}
\min_{\mathbf{x}, \mathbf{y}} & \sum _{k \in \mathcal{K}} y_k & \label{eq:mipobj}\\
s.t. & {\sum_{k \in  \mathcal{K}}x_{i,k} = 1,} & \quad i \in \mathcal{V}, \label{eq:mipcons1}\\
    & \sum _{i\in V}w_{i}x_{i,k}\leq W, & \quad  k \in \mathcal{K}, \label{eq:mipcons2}\\
    & x_{i,k} + x_{j,k} \leq y_{k}, & \quad (i,j) \in \mathcal{E}, k \in \mathcal{K}, \label{eq:mipcons3}\\
    & x_{i,k} \in \{0,1\}, & \quad i \in \mathcal{V}, k \in \mathcal{K}, \label{eq:mipcons4} \\
    & y_{k}\in \{0,1\}, & \quad k \in \mathcal{K}.  \label{eq:mipcons5}
\end{align}
The objective \eqref{eq:mipobj} minimizes the number of bins used; constraint \eqref{eq:mipcons1} ensures that each item must be placed in a bin; constraint \eqref{eq:mipcons2} specifies that the total weight of items placed in a bin must be within its capacity; and constraint \eqref{eq:mipcons3} enforces that the conflict items should not be placed in one bin. This formulation is typically referred to as the compact IP formulation because there is a polynomial number of variables and constraints. The complexity of the problem is closely related to the problem characteristics. As discussed in \cite{BPPC}, existing methods can face significant difficulties when dealing with instances that have a higher bin capacity and/or conflict graphs without showing specific patterns.

It can be seen that the compact IP formulation for BPPC has a symmetric structure. Specifically, a configuration of item placement can correspond to multiple solutions, which can be obtained by shuffling the indices of bins. As a result, it provides a weak LP relaxation bound~\citep{margot2009symmetry}, which can be obtained by relaxing the integer constraints in the compact IP formulation and solving the resulting LP problem. This can significantly slow the branch-and-bound solution process. To address this, Dantzig-Wolfe decomposition can be adopted to obtain an alternative IP formulation of BPPC \cite{vanderbeck2000, vanderbeck2006generic}:  
\begin{align}
\min_{\mathbf{z}} & \sum _{P \in \mathcal{P}} z_P & \label{eq:CGobj}\\
 s.t. & \sum_{P \in \mathcal{P}_i} z_P \geq 1,  & i \in \mathcal{V},\label{eq:CGcons}\\
& z_P \in \{0,1\},   & P \in \mathcal{P}. 
\end{align}
Here, $\mathcal{P}$ denotes the set of all possible patterns to pack items in a bin and $\mathcal{P}_i$ denotes the set of packing patterns that include item $i$. $P$ represents a specific packing pattern associated with a binary decision variable $z_P$, which indicates whether this packing pattern is used. The objective~\eqref{eq:CGobj} is to minimize the number of packing patterns used, and the constraint~\eqref{eq:CGcons} ensures that each item must be covered in at least one of the selected packing patterns. 

This Dantzig-Wolfe reformulation eliminates symmetry by grouping items into packing patterns, and it yields a much stronger LP relaxation bound than the compact IP formulation. However, it typically requires an exponential number of variables (or columns) to represent all packing patterns; therefore, its LP relaxation cannot be solved directly using standard techniques, such as simplex methods. In the next subsection, we introduce CG for solving the LP relaxation of this Dantzig-Wolfe reformulation.

\subsection{Column Generation for BPPC}

To tackle such a large-scale LP, CG takes an iterative procedure, starting from the LP containing a subset of columns (or packing patterns), referred to as the restricted master problem (RMP). In a single iteration, CG solves the RMP and utilize its dual values ($\mathbf{\pi}$) to formulate a pricing problem. The solution to the pricing problem with the most negative reduced costs determines the column that can be added to the RMP. Searching for such columns involves solving a pricing problem, outlined as follows. 
\begin{align}
 \max_{\mathbf{x}} & \sum_{i \in V}\pi_i x_i & \label{eq:ppobj}\\
s.t. & {\sum_{i \in V}w_i x_i\leq W,} & \label{eq:ppcons1}\\
     & x_i + x_j \leq 1, & \quad (i, j) \in \mathcal{E}, \label{eq:ppcons2}\\
     & x_i \in \{0, 1\},  & \quad i \in \mathcal{V}. \label{eq:ppcons3}
\end{align}
Here, the decision variable $x_i$ denotes whether an item $i$ is used to form a packing pattern. The objective function~\eqref{eq:ppobj} aims to minimize the reduced cost ($1 - \sum_{i \in V}\pi_i x_i$), which is equivalent to maximizing $\sum_{i \in V}\pi_i x_i$. Constraints \eqref{eq:ppcons1}-\eqref{eq:ppcons3}  correspond to the Constraints \eqref{eq:mipcons2}-\eqref{eq:mipcons4} in the compact IP formulation, indicating that a feasible solution to the pricing problem must be a valid packing pattern. In fact, the pricing problem is a one-dimensional knapsack problem with conflicts (1DKPC), which is NP-hard~\cite{pferschy2017approximation, pferschy2009knapsack}.

If the minimum reduced cost is negative, a column representing this new packing pattern can be added to the RMP to start the next iteration. Otherwise, the non-negativity of minimum reduced cost indicates that the objective value of the RMP cannot be further improved, and hence the original large LP has been solved to optimality. At this point, the RMP solution is the optimal solution to the original large LP.

The need to repeatedly solve pricing problems is typically a bottleneck in CG~\citep{lubbecke2005selected}. Past studies have explored exact and heuristic pricing methods~\citep{muritiba2010algorithms, hifi2006reactive, enhanceYZ, BPPC}, with extensive empirical studies showing that the performance of CG hinges on multiple evaluation criteria for a pricing method. These criteria can be summarized by the ability to produce a large and diverse set of high-quality columns efficiently. In this regard, ACO, given its ability to generate a diverse set of solutions, can be seen as a promising technique for heuristic pricing introduced in the following.





\subsection{Ant Colony Optimization}
\label{subsec:aco}

ACO is a population-based metaheuristc inspired by biological ants seeking the shortest path between foods and their colony~\citep{as}. It has many applications in large-scale combinatorial optimization~\cite{dorigo2007ant, mavrovouniotis2016ant, xiang2021pairwise, levine2004ant}, such as the classic traveling salesman problem.


ACO maintains a probabilistic distribution that models the likelihood of decision variables taking values of $0$ or $1$ in high-quality solutions, i.e., whether an item is in the 1DKPC solution. It alternates between sampling solutions according to a probabilistic model and updating the models using better-quality solutions, which can be regarded as an online learning model. In the process of constructing a feasible solution, the probabilistic distribution can be defined as follows. 
\begin{equation}\label{equ:aco_p}
        p_{j} = 
    \begin{cases}
        \frac{\tau_{j}^\alpha\eta_{j}^\beta}{\sum_{j\in J}\tau_{j}^\alpha\eta_{j}^\beta}, \;\;\;\;& j \in C \\
        0, & j \notin C.
    \end{cases}
\end{equation}
Here, $C$ represents the set of candidate items that can be selected without violating the constraints of the problem. $\eta_j$ and $\tau_j$, weighted by the hypeparameters $\alpha$ and $\beta$, are critical to the performance of ACO. More specifically, $\eta_j$ is a heuristic measure that can be used to inject prior knowledge about the likelihood that an item $i$ will be used in high-quality solutions. This value is typically set by handcrafted heuristics based on expert knowledge and remains constant throughout the process of ACO. On the other hand, $\tau_j$ denotes the desirability to choose an item $j$ (or the amount of pheromone deposited by ants). It is typically uniformly initialized and updated dynamically as better solutions are sampled, to more accurately indicate the likelihood of the item $j$ shown in high-quality solutions. Intuitively, if an item $j$ is frequently selected in high-quality solutions, the corresponding pheromone value $\tau_j$ is increased to reinforce sampling that variable when constructing new samples.

We introduce a widely used policy to update pheromone values from the ant system~\citep{antsystem}. In each iteration, after constructing a set of $N$ newly sampled solutions, the pheromone values $\tau_i$, where $i = 1, \dots, v $, are updated based on the sample solutions generated: 
\begin{equation}
    \tau_i = (1 - \rho)\tau_i + \sum_{n = 1}^{N} \Delta\tau_{i}^n,
    \label{equ:aco_tau}
\end{equation} 
where $\rho > 0$ is the pheromone evaporation coefficient and $\Delta\tau_{i}^n$ is the amount of pheromone deposited by the $n^{th}$ sample at the item $i$. Let $c_n$ denote the objective value collected by the $n^{th}$ sample, $c_{best}$ represents the best objective value found so far, and $\lambda > 0$ be a constant. We can define  $\Delta\tau_{i}^{n} = c_n/c_{best}/\lambda$, if the item $i$ is selected in the current sampled solution; otherwise $\Delta\tau_{i}^{n} = 0$. The amount of pheromone deposited by an ant when it selects items is proportional to the objective value of the set. As we are solving a maximization problem, items that appear in high quality solutions are reinforced, so that these items are more likely to be selected when constructing solutions in the later iterations.


ACO represents an online learning method which can iteratively improve its optimization performance through sampling using Equation~\eqref{equ:aco_p}. ACO typically starts with an initialization of randomly generated individuals, or individuals injected with simple heuristic rules. However, we can do better by training an offline ML model using data gathered from solved problem instances before ACO's online learning phase. Such a ML-based solution prediction method can significantly accelerate ACO's optimization process.

\subsection{Machine Learning for Optimization}


Machine learning has been shown to be effective in improving combinatorial optimization~\citep{bengio2021machine}, such as learning to make decisions instead of using handcrafted heuristic rules for mixed-integer programming solvers~\citep{zhang2023survey}, metaheuristics~\citep{talbi2021machine, karimi2022machine}, and decomposition techniques~\citep{morabit2021machine, vaclavikAcceleratingBranchandPriceAlgorithm2018, enhanceYZ}. 

Assuming that we have access to data of solved problem instances, then it is possible to employ ML to learn from such data and then generalize it on unseen problem instances. \citet{boosting} proposed MLACO that explores such a solution prediction method to warm-start ACO, substituting either the heuristic weight measure $\eta$ set by some primitive heuristic rule, or the randomly initialized pheromone matrix $\tau$. \citet{ye2023deepaco} proposed a deep neural network to achieve better quality predictions. Another study~\cite{QlearnACO} presents a hybrid ACO method that integrates the problem information extracted from Graphic Neural Network with the standard pheromone and greedy information, employing a Q-learning to learn which type of information is better for solving specific problem instances. Unlike these studies that aim to find a single best solution for standalone optimization problems, we develop a method that combines ML and ACO to generate a diverse set of high-quality columns with the aim to accelerate CG. In particular, the diversity of the generated columns can be crucial to the efficiency of CG, in addition to the solution quality. 

ML has also been used to improve the CG process, such as automatically learning to predict optimal dual values to stabilize CG~\citep{kraul2023machine}, learning column selection rules~\citep{morabit2021machine, chi2022deep}, and learning heuristic pricing methods~\citep{enhanceYZ}. In particular, the MLPH method proposed by \citet{enhanceYZ} can accelerate the CG process by sampling columns based on its ML prediction of the pricing problem. However, the limitation of MLPH is that it samples columns according to a fixed probability distribution proportions to the ML prediction. This work aims to address this limitation by combining ML offline prediction with ACO online learning to further accelerate CG.

\section{The Proposed Approach}
\label{sec:mlaco}


In this section, we describe the proposed heuristic pricing method that is used to address the pricing problem (i.e., 1DKPC) and generate columns at every iteration of the CG. Our hybrid method combines ML and ACO to strike a balance between efficiency, effectiveness, and solution diversity. As shown in Algorithm~\ref{alg:mlaco}, ML is used to make a prediction of the optimal solution to the pricing problem, denoted as $p$, given the features of the decision variables for an unseen pricing problem instance. The ML prediction is then used to accelerate the ACO to quickly find a diverse set of good solutions.

\subsection{Optimal Solution Prediction}\label{subsec:solpred}

We model the solution prediction task for 1DKPC as a binary classification problem, where the output of a decision variable denotes the likelihood of the corresponding item in the optimal solution. In our training set, a training example $(\bm{f}, y)$ corresponds to an item in an optimally solved 1DKPC problem, where $\bm{f}$ represents a set of features that are used to characterize this item and $y$ is its optimal solution value. We describe our four problem-specific features and two statistical features as follows. 

Recall that the 1DKPC problem (Equations~\eqref{eq:ppobj}-\eqref{eq:ppcons3}) aims to find a set of non-conflict items that maximize profit, subject to a capacity limit. The first feature we include is about the profit (i.e., the objective coefficient) of an item $i$ and is normalized by the maximum and minimum profits of items in the same problem instance as follows, 
\begin{equation}
    f_1(i) = \frac{\pi_i - \min_{j \in V}\pi_j}{\max_{j \in V}\pi_j - \min_{j \in V}\pi_j}.
\end{equation}

Our second feature calculates the profit per unit weight ,i.e., the ratio of the profit and weight of item $i$,
\begin{equation}
    f_2(i) = \frac{\pi_i}{w_i}.
\end{equation}
Note that this is a commonly used measure in designing greedy heuristic rules for knapsack problems. Our third feature is about degree of an item and and is normalized by the maximum and minimum profits of items in the same problem instance, 
\begin{equation}
     f_3(i) = \frac{\deg(i) - \min_{j \in V}\deg(j)}{\max_{j \in V}\deg(j) - \min_{j \in V}\deg(j)},
\end{equation}
The fourth feature is an upper bound on the profit if item $i$ is selected, computed by 
\begin{equation}
     f_4(i) = \pi_i + \sum_{(i,j) \notin E}{\pi_j}. 
\end{equation}
It calculates the sum of profits of item $i$ and the items that are not in conflict with item $i$. 


To better characterize decision variables in high-quality solutions, we adopt two statistical features~\cite{usingstatistical} that are computed over a set of $N$ randomly sampled solutions. Our random sampling algorithm is shown in Algorithm~\ref{alg:statsample}. It is obvious that the time complexity to generate one sample of 1DKPC using this method is $\mathcal{O}(|V| + |E|)$ if we represent the conflict graph using an adjacency list, where $|E|$ represents the number of edges in the conflict graph. Hence, the total time complexity of generating $N$ samples is $\mathcal{O}(N(|V| + |E|))$. 

\begin{algorithm}[!tb]
\caption{MLACO for CG}\label{alg:mlaco}
\SetKwInput{KwInput}{Input}                
\SetKwInput{KwOutput}{Output}              
    
    \KwInput{$\mathcal{M}:$ an offline-trained ML model;\newline
            \indent $T:$ iteration number for ACO;\newline
            $(\pi, W, C, V, E):$ problem data;
            }
    \KwOutput{New columns with negative reduced cost}
    Sampling a set of random samples (Alg.~\ref{alg:statsample})\\
    Compute statistical features based on random samples \\
    ML prediction $ p \leftarrow \mathcal{M}(\bm{f})$\\
    Initialize $\eta = p$\\
    Initialize $\tau$ uniformly\\
    \For{t $\leftarrow$  1 to T}{
        Diversity-aware Sampling (Alg.~\ref{alg:sample})\\
        Update $\tau$ according to Eq.~\eqref{equ:aco_tau}\\
    }
\end{algorithm}

\begin{algorithm}[!tb] 
\caption{Random Sampling}\label{alg:statsample}
\SetKwInput{KwInput}{Input}                
\SetKwInput{KwOutput}{Output}              

\DontPrintSemicolon
  \KwInput{$(\pi, W, C, V, E):$ problem data\newline
  $N:$ sample size;}
  \KwOutput{A set of randomly sampled solutions}
    \For{n $\leftarrow$ 1 to N}{
    Initialize a candidate set\\
    \While{candidate set is not empty }
    {
        Randomly select an item from the set of candidates \\
        Update the candidate set\\
    }
}

\end{algorithm}

The first statistical feature measures the correlation between the presence of an item $i$ in the sample solutions and the objective values of the samples.
\begin{equation}
    f_c(i) = \frac{\sum_{n=1}^{N}(s_i^n - \bar{s}_i)(o^n - \bar{o})}{\sqrt{\sum_{n=1}^{N}(s_i^n - \bar{s}_i)^2}\sqrt{\sum_{n=1}^{N}(o^n - \bar{o})^2}},
\end{equation}
where $s_i^n$ is a binary value, indicating whether the item $i$ is a part of the $n^{th}$ sample, $o^n$ represents the objective value of the $n^{th}$ sample, and $\bar{s}_i=\sum_{n=1}^N s_i^n/N$ and $\bar{o} = \sum_{n=1}^N o^n/N$ are the average values across the $N$ samples. An item with a high correlation score indicates that this item is likely to appear in high-quality solutions of the corresponding 1DKPC instance. 

The second statistical measure is based on the ranking of the $N$ sample solutions. Let $r^n$ denote the rank of the $n^{th}$ sample in terms of its objective value in descending order. For an item $i$, this statistical feature accumulates the ranking score across the samples that contain this item: 
\begin{equation}
    f_r(i) = \sum_{n=1}^N \frac{s_i^n}{r^n}.
\end{equation}
An item with a high ranking score indicates that this item appears more frequently in high-quality solutions. 

Given a set of optimally solved 1DKPC instances, we can then construct the training set where each training example corresponds to an item in a 1DKPC instance. For each training example, we extract the six features to characterize the associated item and assign a class label based on whether the item is part of the optimal solution. We train a Support Vector Machine (SVM)~\cite{svm}. to distinguish items that belong to optimal solutions from those that do not.  Given an unseen 1DKPC instance, the trained SVM model can then be used to predict for each item whether it belongs to an optimal solution. More specifically, we can calculate the distance from an item to the decision boundary of SVM in the feature space, which indicates the likelihood that this item belongs to an optimal solution. We employ SVM mainly because of its efficiency in making predictions.

\subsection{MLACO for Column Generation}

Given the offline-trained ML model, we can then use it to make predictions of the optimal solution for unseen problem instances in the iterative process of CG. An accurate prediction can then be used to accelerate ACO to quickly find high-quality solutions. Moreover, we devise diversity-aware sampling to encourage ACO to sample a diverse set of high-quality solutions.

\subsubsection*{Accelerating ACO with ML prediction}
The effectiveness of ACO is heavily dependent on the parameterized probabilistic model, which contains two main parameters $\tau$ and $\eta$. The algorithm automatically learns the value of $\tau$ through searching iterations, while the initialization of $\eta$ typically draws upon prior knowledge provided by experts. In the context of the 1DKPC, the profit-to-weight ratio serves as a robust greedy heuristic rule and is commonly employed in relevant research studies. For each pricing problem, $\eta$ can be set to the profit-to-weight ratio: $\eta_{j} = \pi_i/w_i$, where $\pi$ represents dual solutions retrieved from RMP. This heuristic rule evaluates the item's price per unit weight, making items with higher profit per unit weight are more likely to be selected \cite{coniglio2021new}. We will use this setting for the standard ACO in our experimental comparison. 

In contrast to initialization based on prior knowledge, we set $\eta$ based on our ML predictions, aiming to guide the search toward more promising areas. As mentioned in Section~\ref{subsec:solpred}, we train a binary SVM model to predict whether an item belongs to the optimal solution for a pricing problem, that is, 1DKPC. We use Platt scaling to transform the SVM prediction from the class label to a probability distribution over classes \cite{platt1999probabilistic}. The transformation produces probability estimates given by 
\begin{equation}
{P}(y=1|x)=\frac  {1}{1+\exp(af(x)+b)},
\end{equation}
where $a$ and $b$ are two learned parameters. This expression represents a logistic transformation of the classifier scores $f(x)$ into a probability value that indicates the probability that this item belongs to an optimal solution. We use the probability values to set $\eta$ in the ACO algorithm by default and will also explore and discuss other possible uses of the ML predictions to improve ACO in our experimental study.

\subsubsection*{Diversity-aware Sampling}\label{subsec:sampling}



\begin{algorithm}[!tb]
\caption{Diversity-aware Sampling}\label{alg:sample}
\SetKwInput{KwInput}{Input}                
\SetKwInput{KwOutput}{Output}              
  \KwInput{$(\pi, W, C, V, E):$ problem data\newline
            $p:$ probabilistic model}
  \KwOutput{Sampled columns with negative reduced costs}
    \SetKwRepeat{Do}{do}{while}
    \For{each item}{
        Add this initial item to the working solution.\\
        \Do{the candidate set of items is not empty}{Update candidate set \\
        Update the probabilistic model according to Eq.~\ref{equ:aco_p}\\
        Sample an item from candidate set; add this item to the working solution\\
        }
    }
\end{algorithm}

In CG, sampling a diverse set of columns can be crucial to its performance. To better align this objective, we propose a diversity-aware sampling method to replace the traditional sampling method in ACO. We generate columns based on the ML prediction while ensuring that each item is covered at least once, i.e., at each iteration, assuming that the sample size is equal to the number of items, we initiate the sampling process by starting from the first item and progressing to the last. This ensures that each item is treated as the first added item at least once. As shown in Algorithm~\ref{alg:sample}, for every sampled solution, once the first item $k_1$  is selected, we choose the remaining items based on the ML prediction. More specifically, our method to generate one column includes the following steps:
\begin{itemize}[nolistsep]
    \item Initialize an item with the starting item $k_1$.
    \item Generate candidate items $k_j$ that can be visited.
    \item Select the next item to be included in the sampled column that does not violate the capacity and conflict constraints.
\end{itemize}
Note that we only include columns with negative reduced cost into RMP in each sampling iteration.



\section{Empirical Studies}
\subsection{Experimental Setup}

\subsubsection*{Bin Packing Benchmarks and Conflict Graph Generation}

We consider a standard bin packing benchmark, Hard28~\citep{Hard28}, for testing. Hard28 consists of $28$ problem instances with a uniform bin capacity $1000$. For a problem instance, the number of items is in $\{160, 180, 200\}$. Following previous work~\citep{BPPC}, we model the conflicts between items using randomly generated graphs with density values around $0.5$. We also consider different capacity multipliers, $\{1,2,5,15\}$. These settings can result in more challenging pricing problems as discussed in~\citep{BPPC}. For training, we use a set of $20$ small problem instances with $120$ items from \citet{falkenauer1996hybrid}, different from the test problem instances. Each of these training problem instances has 120 items, and their weights are uniformly distributed.

\subsubsection*{Data collection and training}
We use CG to solve a training problem instance. In this process, multiple pricing problems are solved to optimality, and we record the optimal solution values and the features of pricing problem variables to form the training data. Specifically, we consider pricing problems in every five iterations to increase the diversity of training data. Moreover, we start the recording from the $10^{th}$ iteration and continue to the $30^{th}$ iteration of CG, because we observe that these pricing problem instances are relatively easy to solve in the early iterations. The statistical features are calculated from a set of $N$ randomly sampled solutions, where $N$ is the number of items. All features are normalized on an instance-wise basis.

With the training data, we train a linear SVM using the standard software package~\cite{libSVM}. We configure the SVM to directly output the probabilistic values between $0$ and $1$, and set the regularization term according to the ratio between the negative training examples and the positive ones. The regularization term controls the importance of correctly classifying positive and negative training examples, and our setting of the regularization parameter specifies that these two situations are equally important. We set the remaining hyperparameters to the default values. 

\subsubsection*{Compared Methods} The following methods are compared in our empirical study: 1) \textbf{MLACO} (the proposed method)\footnote{Our source code is written in C++ and is available at \url{https://github.com/hongjie-ml/MLACO\_binpack}}: We use MLACO to solve a pricing problem and add all newly generated columns with negative reduced costs to the RMP to start the next iteration of CG. The heuristic value $\eta$ is set to the ML prediction, and the pheromone value $\tau$ is initialized uniformly. We set $\alpha$ and $\beta$ to $1$, and set $\rho$ to $0.95$. The iteration number is set to $10$ and the population size is set to the number of items in a problem instance; 2) \textbf{Gurobi$^s$}: We use the Gurobi solver~\cite{gurobi} to solve the pricing problem to optimality and add the optimal column to start the next iteration of CG. Note that Gurobi is a state-of-the-art mixed-integer programming solver; 3) \textbf{Gurobi$^m$}: In this setting, we use Gurobi to solve a pricing problem to optimality and add multiple columns with negative reduced costs in addition to the optimal column. We turn on the solution pool feature of Gurobi to encourage it to find multiple high-quality solutions; 4) \textbf{MLPH} \citep{enhanceYZ}: A method that samples columns based on ML prediction of the optimal solution on the  graph coloring problem. Compared to our method, the probabilistic model in MLPH is set according to the ML prediction and remains fixed throughout the sampling process. 5) \textbf{ACO}: A widely-used metaheuristic for combinatorial optimization as introduced in Section~\ref{subsec:aco}. The parameters in ACO are the same as MLACO. We set the heuristic value $\eta$ to a widely-used heuristic rule, the profit-to-weight ratio.
    
    

    

We initialize the RMP with a set of randomly generated columns such that the RMP is feasible, to start the solution process of CG. The pricing problem in an iteration of CG is solved by a compared method. For MLACO, MLPH, and ACO, they are heuristic pricing methods without the guarantee of finding the optimal solution. When these methods do not find any column with a negative reduced cost, we execute the exact method Gurobi$^s$ to find the optimal column. If the minimum reduced cost is negative, CG proceeds to the next iteration. Otherwise, CG reaches optimality and ends. 

In the remainder of this section, we first compare the efficacy of different pricing methods in terms of the performance of CG to solve the LP relaxations for the BPPC. Then, we report the results for branch-and-price for solving the integer problem BPPC, where CG with a pricing method is used to derive LP relaxation bounds at every node of the branch-and-bound tree. In all experimental settings, CG is initialized using a set of randomly generated columns and is subject to a cutoff time of $1800$ seconds.

\subsection{Results for CG}



\begin{figure}[!tb]
    \centering
    \includegraphics[scale=0.45]{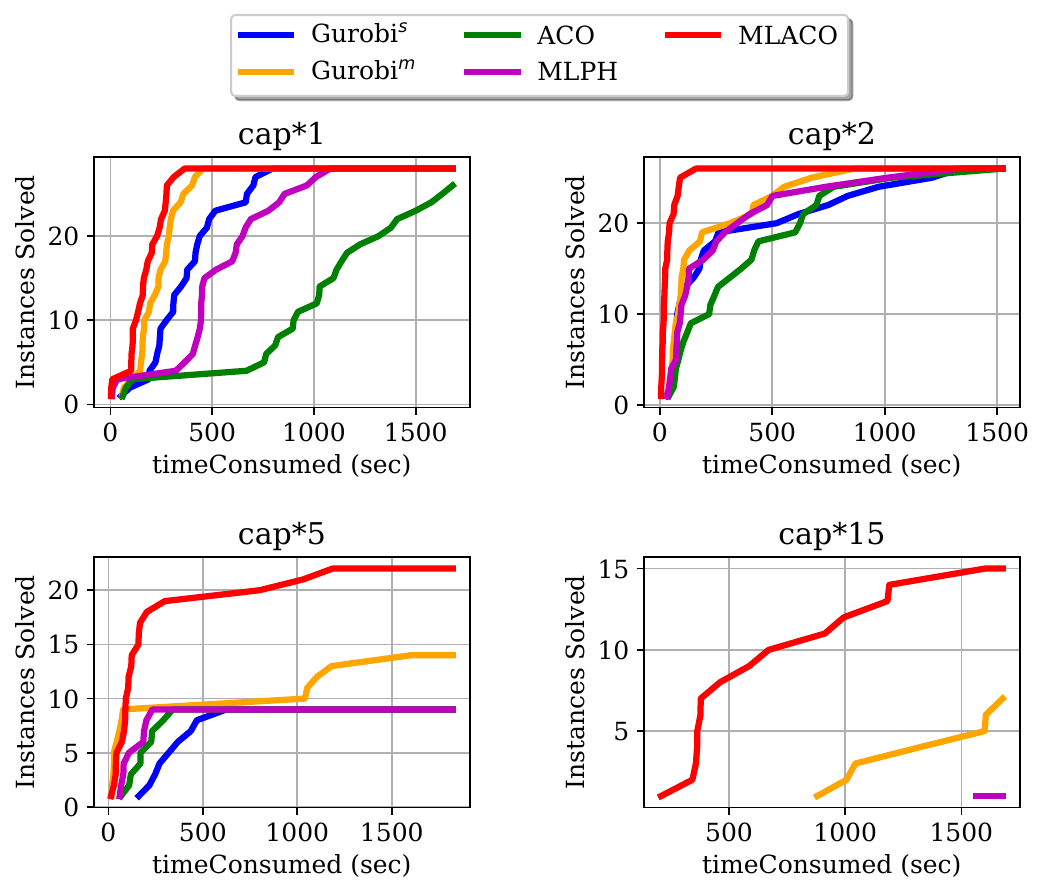}
    \caption{Number of instances solved by CG using different pricing methods. Each subfigure shows the results for a different bin capacity multipler.}
    \label{fig:timevsnumbersolved}
\end{figure}

Figure~\ref{fig:timevsnumbersolved} shows the number of solved problem instances for a method within a certain time limit. For instance, under the capacity multiplier $15$, in the first $1,500$ seconds, our MLACO method solves $15$ LP instances, Gurobi$^m$ solves $5$ LP instances, and the rest of the methods do not solve any. Overall, we can observe that MLACO can solve more problem instances at a given time limit under all capacity settings. The performance gap between MLACO and the compared methods increases as the bin capacity increases, and this is because the higher bin capacity leads to more difficult pricing problems. Specifically, we can first observe that MLACO outperforms ACO by a large margin. This shows the benefit of incorporating ML techniques over handcrafted heuristics. Our proposed MLACO also outperforms MLPH significantly. This shows the efficacy of updating the probabilistic model during the sampling process, resulting in more efficient sampling of high-quality solutions. It can also be seen that Gurobi$^m$ better accelerates CG compared to Gurobi$^s$, showing the benefits of generating multiple high-quality columns in an iteration of CG. Similar observations can be made in Table~\ref{tab:1}, which shows the number of optimally solved problem instances at the cutoff time. Most noticeably, MLACO still manages to solve $15$ LP instances when the capacity multiplier equals $15$, demonstrating its capability under more difficult pricing problem settings.

\begin{table}[!tb]
    \centering
\begin{tabular}{lrrrr}
\toprule
Method & cap*1 & cap*2 & cap*5 & cap*15 \\
\midrule
Gurobi$^s$ & 28 & 26 & 9 & 0 \\
Gurobi$^m$ & 28 & 26 & 14 & 7 \\
ACO & 26 & 26 & 9 & 1 \\
MLPH & 28 & 26 & 10 & 1 \\
MLACO & 28 & 26 & \textbf{23} & \textbf{15} \\
\bottomrule
\end{tabular}
    \caption {Number of instances solved by CG with various pricing methods for different capacity multipliers.}
    \label{tab:1}
\end{table}


In addition, we present the runtimes of CG with various pricing methods averaged across all problem instances in Table~\ref{tab:cgaverage}. Note that if a method cannot solve a problem instance within the cutoff time, the runtime is set to the cutoff time $1800$, which is used to compute the average results. We can observe that CG with our MLACO method achieves significantly shorter runtimes compared to other methods, and this observation is consistent across problem instances with all different capacity multipliers.


\begin{table}[!tb]
    \centering
\begin{tabular}{lllll}
\toprule
Method & cap*1 & cap*2 & cap*5 & cap*15 \\
\midrule
Gurobi$^s$ & 383.26 & 426.65 & 1332.62 & 1800.00 \\
Gurobi$^m$ & 236.75 & 320.77 & 1126.82 & 1676.12 \\
ACO & 1059.76 & 494.99 & 1283.05 & 1792.43 \\
MLPH & 535.66 & 379.06 & 1264.46 & 1791.52 \\
MLACO & \textbf{171.31} & \textbf{162.43} & \textbf{566.23} & \textbf{1192.31} \\
\bottomrule
\end{tabular}
    \caption{Solving time of CG with different pricing methods averaged across all problem instances.}
    \label{tab:cgaverage}
\end{table}

\subsubsection*{Ablation Study} Table~\ref{tab:samplingablation} presents the results of an ablation study on several different ways of combining ML and ACO. Specifically, we examine the following three variants:
\begin{itemize}[nolistsep]
    \item {\bf mlaco\_predicted\_eta}: Set the $\eta$ value to the ML prediction value and initialize $\tau$ uniformly, as that in Section~\ref{sec:mlaco};
    \item {\bf mlaco\_pred\_heu\_eta}: Set the $\eta$ value to the product of the predicted probability and profit-to-weight ratio: $\eta_i = p_i \cdot \pi_i/ w_i$ and initialize $\tau$ uniformly;
    \item {\bf mlaco\_predicted\_tau}: Set the $\tau$ value to the ML prediction value and initialize $\eta$ uniformly.
\end{itemize}

\begin{table}[!tb]
    \centering
    \resizebox{0.48\textwidth}{!}{
    \begin{tabular}{lrrrrr}
    \toprule
    Method & Sampling & cap*1 & cap*2 & cap*5 & cap*15 \\
    \midrule
    mlaco\_predicted\_eta & Y & 28 & 26 & 23 & 15 \\
    mlaco\_pred\_heu\_eta & Y & 28 & 26 & 13 & 5 \\
    mlaco\_predicted\_tau & Y & 28 & 26 & 18 & 14 \\
    \midrule
    mlaco\_predicted\_eta& N & 27 & 24 & 17 & 12 \\
    mlaco\_pred\_heu\_eta & N & 27 & 25 & 9 & 2 \\
    mlaco\_predicted\_tau& N & 25 & 26 & 16 & 12 \\
    \bottomrule
    \end{tabular}
    }
    \caption{Comparison of MLACO variants with and without the use of diversity-aware sampling.}
    \label{tab:samplingablation}
\end{table}

We can observe that initializing $\eta$ with the ML prediction value achieves the best performance across all three model variants. This is closely followed by initializing $\tau$ with the ML prediction value. The variant which sets the value of $\eta$ to the product of the predicted probability and a greedy heuristic rule performs worse when the capacity multiplier becomes larger. One possible explanation is that the greedy heuristic rule biases the search in a wrong direction, as ACO lacks information about the conflict graph. 

Finally, we examine the efficacy of the proposed diversity-aware sampling method compared to the regular sampling method in ACO. Recall that our method encourages sample diversity by ensuring that each item must be covered at least once by the newly generated columns. The results of MLACO without using diversity-aware sampling are shown in the bottom half of Table~\ref{tab:samplingablation}. We can see that diverse-aware sampling significantly improves the performance of MLACO variants across all capacity settings. This result shows the importance of generating a diverse set of columns for CG.


\subsection{Results for Branch-and-Price}
Our experiments have shown that the proposed MLACO can efficiently generate a diverse set of high-quality columns, thereby accelerating CG to derive tight LP relaxation bounds for BPPC. In this part, we leverage CG with our pricing method for enhancing branch-and-bound, to obtain an optimal integer solution to BPPC. Branch-and-bound is a canonical method for exact combinatorial optimization. It works by decomposing the integer problem into smaller subproblems and exploring the subproblems recursively. This process can be seen as a tree search, where each node represents a subproblem. A critical component of branch-and-bound is its bounding function, which determines whether a node can be safely pruned without a further visit. Specifically, a node can be pruned if its best possible solution (e.g., the LP bound) is no better than the objective value of the incumbent solution. Here, we use CG at each node to produce tight LP bounds. Note that branch-and-bound with CG, commonly called branch-and-price, is very effective on a range of problems~\citep{barnhart1998branch}. 

Our branch-and-price implementation is based on an academic mixed-integer programming solver, SCIP~\cite{GleixnerEtal2018OO}. Most importantly, we adopt the Ryan/Foster branching~\citep{ryan1981integer}, which is commonly preferred in branch-and-price. Specifically, a problem is decomposed into two subproblems, by adding constraints that specify that a pair of items that are either a) forced to be packed together or b) prohibited from being packed together. Note that this branching rule can play a crucial role in the solution time of branch-and-price, in addition to the bounding function. 



\begin{figure}[!tb]
    \centering
    \includegraphics[scale=0.45]{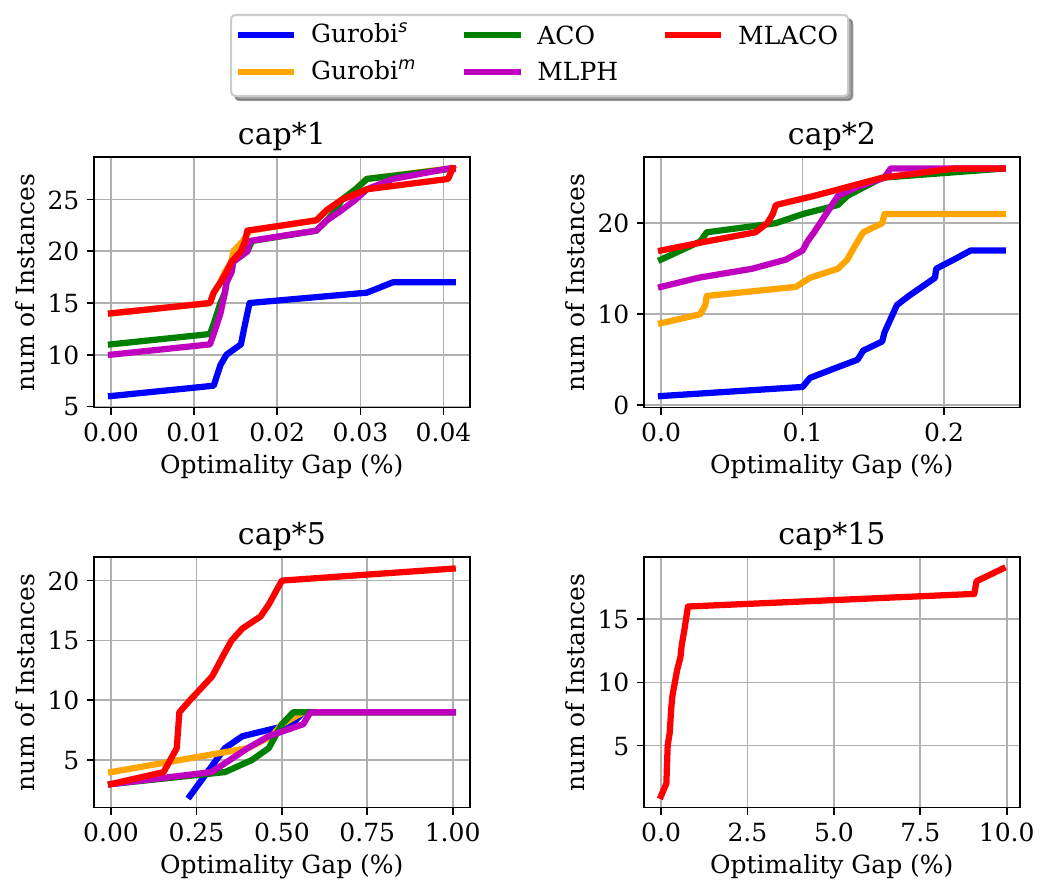}
    \caption{Number of instances solved within an optimality gap by branch-and-price with various pricing methods.}
    \label{fig:BPgapNumsolved}
\end{figure}

Figure~\ref{fig:BPgapNumsolved} shows the number of instances solved within a particular optimality gap by the branch-and-price algorithm with different pricing methods. The optimality gap measures the difference between the objective value of the best-found solution and the worst LP objective among all leaf nodes in the branch-and-bound tree. The optimality gap reduces to $0$ when a problem is solved to optimality. It can be seen that MLACO achieves very competitive performance under small capacity multipliers $1$ and $2$. With high capacity multipliers $5$ and $15$, branch-and-price with MLACO solves much more problem instances within a certain optimality gap. For instance, with the capacity multiplier $5$, MLACO solves $20$ out of $28$ instances to less than 0.5\% of the optimality gap, while the compared methods achieve this result for less than $10$ problem instances. The main reason for such a significant difference in performance is that our pricing heuristic, MLACO, can still solve many LP relaxations at the root node under large capacity multiplier settings. In contrast, the compared methods struggle to solve the root LP; therefore, the optimality gap does not exist. Table~\ref{tab:BPsolved} reports the number of problem instances optimally solved by each method (i.e., the optimality gap is $0$) under different capacity multipliers.

In Table~\ref{tab:bp_time}, we report the solution time of branch-and-price with various pricing methods for the problem instances with default bin capacity (multiplier equal to $1$). MLACO obtains the shortest solution time for $10$ problem instances, and the compared methods achieve the shortest solution time in $8$ problem instances in total. Specifically, MLACO outperforms ACO and MLPH in a majority of the problem instances, because it inherits the benefits from both offline learning (i.e., ML) and online learning (i.e., ACO). It can be seen that the performance of MLACO and Gurobi$^m$ tends to differ substantially among the test problem instances. This should be understandable because Gurobi is based on mixed-integer-programming techniques and is very different from the nature of MLACO.

\begin{table}[!tb]
\centering
\begin{tabular}{lrrrr}
\toprule
Method & cap*1 & cap*2 & cap*5 & cap*15 \\
\midrule
Gurobi$^s$ & 6 & 1 & 0 & 0 \\
Gurobi$^m$ & \textbf{14} & 9 & \textbf{4} & 0 \\
ACO & 11 & 16 & 3 & 0 \\
MLPH & 10 & 13 & 3 & 0 \\
MLACO & \textbf{14} & \textbf{17} & 3 & \textbf{1} \\
\bottomrule
\end{tabular}
\caption{Number of instances solved to optimality by branch-and-price with various pricing methods.}
\label{tab:BPsolved}
\end{table}

\begin{table}[tb!]
    \centering
\begin{tabular}{lrrrrr}
\toprule
Instance & Gurobi$^s$ & Gurobi$^m$ & ACO & MLPH & MLACO \\
\midrule
BPP832 & - & \textbf{1730.8} & - & - & - \\
BPP485 & - & \textbf{1398.3} & - & - & - \\
BPP181 & 1331.4 & \textbf{518.7} & - & - & 945.8 \\
\midrule
BPP47 & 648.5 & 243.7 & \textbf{78.6} & 312.8 & 127.3 \\
BPP640 & 1112.7 & 606.1 & \textbf{44.5} & 133.9 & 64.0 \\
\midrule
BPP60 & - & 1531.3 & 1745.9 & \textbf{1143.3} & - \\
BPP14 & - & 1542.7 & 1018.2 & \textbf{1009.3} & - \\
BPP814 & 1138.8 & 758.9 & 1627.9 & \textbf{374.2} & 679.4 \\
\midrule
BPP13 & - & - & - & - & \textbf{1441.5} \\
BPP709 & - & - & - & - & \textbf{1320.2} \\
BPP40 & - & - & 1654.2 & - & \textbf{1223.6} \\
BPP766 & - & - & 1527.2 & 1463.7 & \textbf{1169.2} \\
BPP645 & - & 1714.5 & - & - & \textbf{892.1} \\
BPP716 & - & 1565.7 & - & - & \textbf{674.6} \\
BPP531 & - & 798.3 & 583.8 & 532.3 & \textbf{432.1} \\\
BPP360 & 832.7 & 961.6 & 539.8 & 496.0 & \textbf{411.8} \\
BPP359 & - & 699.7 & 524.9 & 610.3 & \textbf{351.3} \\
BPP742 & 1646.5 & 620.0 & 371.5 & 581.9 & \textbf{118.5} \\

\bottomrule
\end{tabular}
    \caption{Solution time for problem instances solved by at least one method. `-' denotes that a method does not solve the problem instance within the cutoff time.}
    \label{tab:bp_time}
\end{table}

\section{Conclusion}

In this paper, we introduced a hybrid approach called MLACO that combines machine learning (ML) with ant colony optimization (ACO) to boost Column Generation (CG). We trained an offline ML model based on historical data to predict the optimal solution to a pricing problem in CG and integrated the ML prediction with the probabilistic model of ACO to generate multiple high-quality columns. We explored different ways of combining ML predictions with ACO and proposed a diversity-aware sampling method to improve the diversity of generated columns, which is crucial for CG. We evaluated our method on the bin packing problem with conflicts and showed that our MLACO method significantly improved the performance of CG compared to several state-of-the-art methods, especially when pricing problems are difficult. We also showed that our method significantly reduced the solution time of a Branch-and-Price algorithm for generating optimal integer solutions.

There are several potential avenues for further research. First, our current method relies on manual feature extraction, and thus using graph neural networks could offer a promising strategy for automatic feature extraction. Second, it would be interesting to extend our method to solve other combinatorial optimization problems, such as vehicle routing problems, where pricing problems are shortest-path problems. Finally, exploring the use of reinforcement learning to learn an intelligent policy for generating columns in solving pricing problems would be another interesting direction for future investigation.


\bibliographystyle{ACM-Reference-Format}
\bibliography{reference.bib}


\begin{thebibliography}{39}


\ifx \showCODEN    \undefined \def \showCODEN     #1{\unskip}     \fi
\ifx \showDOI      \undefined \def \showDOI       #1{#1}\fi
\ifx \showISBNx    \undefined \def \showISBNx     #1{\unskip}     \fi
\ifx \showISBNxiii \undefined \def \showISBNxiii  #1{\unskip}     \fi
\ifx \showISSN     \undefined \def \showISSN      #1{\unskip}     \fi
\ifx \showLCCN     \undefined \def \showLCCN      #1{\unskip}     \fi
\ifx \shownote     \undefined \def \shownote      #1{#1}          \fi
\ifx \showarticletitle \undefined \def \showarticletitle #1{#1}   \fi
\ifx \showURL      \undefined \def \showURL       {\relax}        \fi
\providecommand\bibfield[2]{#2}
\providecommand\bibinfo[2]{#2}
\providecommand\natexlab[1]{#1}
\providecommand\showeprint[2][]{arXiv:#2}

\bibitem[Barnhart et~al\mbox{.}(1998)]%
        {barnhart1998branch}
\bibfield{author}{\bibinfo{person}{Cynthia Barnhart}, \bibinfo{person}{Ellis~L Johnson}, \bibinfo{person}{George~L Nemhauser}, \bibinfo{person}{Martin~WP Savelsbergh}, {and} \bibinfo{person}{Pamela~H Vance}.} \bibinfo{year}{1998}\natexlab{}.
\newblock \showarticletitle{Branch-and-price: Column generation for solving huge integer programs}.
\newblock \bibinfo{journal}{\emph{Operations research}} \bibinfo{volume}{46}, \bibinfo{number}{3} (\bibinfo{year}{1998}), \bibinfo{pages}{316--329}.
\newblock


\bibitem[Bengio et~al\mbox{.}(2021)]%
        {bengio2021machine}
\bibfield{author}{\bibinfo{person}{Yoshua Bengio}, \bibinfo{person}{Andrea Lodi}, {and} \bibinfo{person}{Antoine Prouvost}.} \bibinfo{year}{2021}\natexlab{}.
\newblock \showarticletitle{Machine learning for combinatorial optimization: a methodological tour d’horizon}.
\newblock \bibinfo{journal}{\emph{European Journal of Operational Research}} \bibinfo{volume}{290}, \bibinfo{number}{2} (\bibinfo{year}{2021}), \bibinfo{pages}{405--421}.
\newblock


\bibitem[Boser et~al\mbox{.}(1992)]%
        {svm}
\bibfield{author}{\bibinfo{person}{Bernhard~E. Boser}, \bibinfo{person}{Isabelle~M. Guyon}, {and} \bibinfo{person}{Vladimir~N. Vapnik}.} \bibinfo{year}{1992}\natexlab{}.
\newblock \showarticletitle{A training algorithm for optimal margin classifiers}. In \bibinfo{booktitle}{\emph{Proceedings of the Fifth Annual Workshop on Computational Learning Theory}} (Pittsburgh, Pennsylvania, USA) \emph{(\bibinfo{series}{COLT '92})}. \bibinfo{publisher}{Association for Computing Machinery}, \bibinfo{address}{New York, NY, USA}, \bibinfo{pages}{144–152}.
\newblock
\showISBNx{089791497X}
\urldef\tempurl%
\url{https://doi.org/10.1145/130385.130401}
\showDOI{\tempurl}


\bibitem[Chang and Lin(2011)]%
        {libSVM}
\bibfield{author}{\bibinfo{person}{Chih-Chung Chang} {and} \bibinfo{person}{Chih-Jen Lin}.} \bibinfo{year}{2011}\natexlab{}.
\newblock \showarticletitle{{LIBSVM}: A library for support vector machines}.
\newblock \bibinfo{journal}{\emph{ACM Transactions on Intelligent Systems and Technology}}  \bibinfo{volume}{2} (\bibinfo{year}{2011}), \bibinfo{pages}{27:1--27:27}.
\newblock
Issue 3.
\newblock
\shownote{Software available at \url{http://www.csie.ntu.edu.tw/~cjlin/libsvm}}.


\bibitem[Chi et~al\mbox{.}(2022)]%
        {chi2022deep}
\bibfield{author}{\bibinfo{person}{Cheng Chi}, \bibinfo{person}{Amine Aboussalah}, \bibinfo{person}{Elias Khalil}, \bibinfo{person}{Juyoung Wang}, {and} \bibinfo{person}{Zoha Sherkat-Masoumi}.} \bibinfo{year}{2022}\natexlab{}.
\newblock \showarticletitle{A deep reinforcement learning framework for column generation}.
\newblock \bibinfo{journal}{\emph{Advances in Neural Information Processing Systems}}  \bibinfo{volume}{35} (\bibinfo{year}{2022}), \bibinfo{pages}{9633--9644}.
\newblock


\bibitem[Coniglio et~al\mbox{.}(2021)]%
        {coniglio2021new}
\bibfield{author}{\bibinfo{person}{Stefano Coniglio}, \bibinfo{person}{Fabio Furini}, {and} \bibinfo{person}{Pablo San~Segundo}.} \bibinfo{year}{2021}\natexlab{}.
\newblock \showarticletitle{A new combinatorial branch-and-bound algorithm for the knapsack problem with conflicts}.
\newblock \bibinfo{journal}{\emph{European Journal of Operational Research}} \bibinfo{volume}{289}, \bibinfo{number}{2} (\bibinfo{year}{2021}), \bibinfo{pages}{435--455}.
\newblock


\bibitem[Dorigo(2007)]%
        {dorigo2007ant}
\bibfield{author}{\bibinfo{person}{Marco Dorigo}.} \bibinfo{year}{2007}\natexlab{}.
\newblock \showarticletitle{Ant colony optimization}.
\newblock \bibinfo{journal}{\emph{Scholarpedia}} \bibinfo{volume}{2}, \bibinfo{number}{3} (\bibinfo{year}{2007}), \bibinfo{pages}{1461}.
\newblock


\bibitem[Dorigo et~al\mbox{.}(1996a)]%
        {as}
\bibfield{author}{\bibinfo{person}{M. Dorigo}, \bibinfo{person}{V. Maniezzo}, {and} \bibinfo{person}{A. Colorni}.} \bibinfo{year}{1996}\natexlab{a}.
\newblock \showarticletitle{Ant system: optimization by a colony of cooperating agents}.
\newblock \bibinfo{journal}{\emph{IEEE Transactions on Systems, Man, and Cybernetics, Part B (Cybernetics)}} \bibinfo{volume}{26}, \bibinfo{number}{1} (\bibinfo{year}{1996}), \bibinfo{pages}{29--41}.
\newblock
\urldef\tempurl%
\url{https://doi.org/10.1109/3477.484436}
\showDOI{\tempurl}


\bibitem[Dorigo et~al\mbox{.}(1996b)]%
        {antsystem}
\bibfield{author}{\bibinfo{person}{M. Dorigo}, \bibinfo{person}{V. Maniezzo}, {and} \bibinfo{person}{A. Colorni}.} \bibinfo{year}{1996}\natexlab{b}.
\newblock \showarticletitle{Ant system: optimization by a colony of cooperating agents}.
\newblock \bibinfo{journal}{\emph{IEEE Transactions on Systems, Man, and Cybernetics, Part B (Cybernetics)}} \bibinfo{volume}{26}, \bibinfo{number}{1} (\bibinfo{year}{1996}), \bibinfo{pages}{29--41}.
\newblock
\urldef\tempurl%
\url{https://doi.org/10.1109/3477.484436}
\showDOI{\tempurl}


\bibitem[Falkenauer(1996)]%
        {falkenauer1996hybrid}
\bibfield{author}{\bibinfo{person}{Emanuel Falkenauer}.} \bibinfo{year}{1996}\natexlab{}.
\newblock \showarticletitle{A hybrid grouping genetic algorithm for bin packing}.
\newblock \bibinfo{journal}{\emph{Journal of heuristics}}  \bibinfo{volume}{2} (\bibinfo{year}{1996}), \bibinfo{pages}{5--30}.
\newblock


\bibitem[Gleixner et~al\mbox{.}(2018)]%
        {GleixnerEtal2018OO}
\bibfield{author}{\bibinfo{person}{Ambros Gleixner}, \bibinfo{person}{Michael Bastubbe}, \bibinfo{person}{Leon Eifler}, \bibinfo{person}{Tristan Gally}, \bibinfo{person}{Gerald Gamrath}, \bibinfo{person}{Robert~Lion Gottwald}, \bibinfo{person}{Gregor Hendel}, \bibinfo{person}{Christopher Hojny}, \bibinfo{person}{Thorsten Koch}, \bibinfo{person}{Marco~E. L{\"u}bbecke}, \bibinfo{person}{Stephen~J. Maher}, \bibinfo{person}{Matthias Miltenberger}, \bibinfo{person}{Benjamin M{\"u}ller}, \bibinfo{person}{Marc~E. Pfetsch}, \bibinfo{person}{Christian Puchert}, \bibinfo{person}{Daniel Rehfeldt}, \bibinfo{person}{Franziska Schl{\"o}sser}, \bibinfo{person}{Christoph Schubert}, \bibinfo{person}{Felipe Serrano}, \bibinfo{person}{Yuji Shinano}, \bibinfo{person}{Jan~Merlin Viernickel}, \bibinfo{person}{Matthias Walter}, \bibinfo{person}{Fabian Wegscheider}, \bibinfo{person}{Jonas~T. Witt}, {and} \bibinfo{person}{Jakob Witzig}.} \bibinfo{year}{2018}\natexlab{}.
\newblock \bibinfo{booktitle}{\emph{The {{SCIP}} Optimization Suite 6.0}}.
\newblock \bibinfo{type}{Technical Report}. \bibinfo{institution}{{Optimization Online}}.
\newblock


\bibitem[{Gurobi Optimization, LLC}(2023)]%
        {gurobi}
\bibfield{author}{\bibinfo{person}{{Gurobi Optimization, LLC}}.} \bibinfo{year}{2023}\natexlab{}.
\newblock \bibinfo{title}{{Gurobi Optimizer Reference Manual}}.
\newblock
\newblock
\urldef\tempurl%
\url{https://www.gurobi.com}
\showURL{%
\tempurl}


\bibitem[Hifi and Michrafy(2006)]%
        {hifi2006reactive}
\bibfield{author}{\bibinfo{person}{Mhand Hifi} {and} \bibinfo{person}{Mustapha Michrafy}.} \bibinfo{year}{2006}\natexlab{}.
\newblock \showarticletitle{A reactive local search-based algorithm for the disjunctively constrained knapsack problem}.
\newblock \bibinfo{journal}{\emph{Journal of the Operational Research Society}} \bibinfo{volume}{57}, \bibinfo{number}{6} (\bibinfo{year}{2006}), \bibinfo{pages}{718--726}.
\newblock


\bibitem[Karimi-Mamaghan et~al\mbox{.}(2022)]%
        {karimi2022machine}
\bibfield{author}{\bibinfo{person}{Maryam Karimi-Mamaghan}, \bibinfo{person}{Mehrdad Mohammadi}, \bibinfo{person}{Patrick Meyer}, \bibinfo{person}{Amir~Mohammad Karimi-Mamaghan}, {and} \bibinfo{person}{El-Ghazali Talbi}.} \bibinfo{year}{2022}\natexlab{}.
\newblock \showarticletitle{Machine learning at the service of meta-heuristics for solving combinatorial optimization problems: A state-of-the-art}.
\newblock \bibinfo{journal}{\emph{European Journal of Operational Research}} \bibinfo{volume}{296}, \bibinfo{number}{2} (\bibinfo{year}{2022}), \bibinfo{pages}{393--422}.
\newblock


\bibitem[Kraul et~al\mbox{.}(2023)]%
        {kraul2023machine}
\bibfield{author}{\bibinfo{person}{Sebastian Kraul}, \bibinfo{person}{Markus Seizinger}, {and} \bibinfo{person}{Jens~O Brunner}.} \bibinfo{year}{2023}\natexlab{}.
\newblock \showarticletitle{Machine learning--supported prediction of dual variables for the cutting stock problem with an application in stabilized column generation}.
\newblock \bibinfo{journal}{\emph{INFORMS Journal on Computing}} (\bibinfo{year}{2023}).
\newblock


\bibitem[Levine and Ducatelle(2004)]%
        {levine2004ant}
\bibfield{author}{\bibinfo{person}{John Levine} {and} \bibinfo{person}{Frederick Ducatelle}.} \bibinfo{year}{2004}\natexlab{}.
\newblock \showarticletitle{Ant colony optimization and local search for bin packing and cutting stock problems}.
\newblock \bibinfo{journal}{\emph{Journal of the Operational Research society}} \bibinfo{volume}{55}, \bibinfo{number}{7} (\bibinfo{year}{2004}), \bibinfo{pages}{705--716}.
\newblock


\bibitem[L{\"u}bbecke and Desrosiers(2005)]%
        {lubbecke2005selected}
\bibfield{author}{\bibinfo{person}{Marco~E L{\"u}bbecke} {and} \bibinfo{person}{Jacques Desrosiers}.} \bibinfo{year}{2005}\natexlab{}.
\newblock \showarticletitle{Selected topics in column generation}.
\newblock \bibinfo{journal}{\emph{Operations research}} \bibinfo{volume}{53}, \bibinfo{number}{6} (\bibinfo{year}{2005}), \bibinfo{pages}{1007--1023}.
\newblock


\bibitem[Margot(2009)]%
        {margot2009symmetry}
\bibfield{author}{\bibinfo{person}{Fran{\c{c}}ois Margot}.} \bibinfo{year}{2009}\natexlab{}.
\newblock \showarticletitle{Symmetry in integer linear programming}.
\newblock \bibinfo{journal}{\emph{50 Years of Integer Programming 1958-2008: From the Early Years to the State-of-the-Art}} (\bibinfo{year}{2009}), \bibinfo{pages}{647--686}.
\newblock


\bibitem[Mavrovouniotis et~al\mbox{.}(2016)]%
        {mavrovouniotis2016ant}
\bibfield{author}{\bibinfo{person}{Michalis Mavrovouniotis}, \bibinfo{person}{Felipe~M M{\"u}ller}, {and} \bibinfo{person}{Shengxiang Yang}.} \bibinfo{year}{2016}\natexlab{}.
\newblock \showarticletitle{Ant colony optimization with local search for dynamic traveling salesman problems}.
\newblock \bibinfo{journal}{\emph{IEEE transactions on cybernetics}} \bibinfo{volume}{47}, \bibinfo{number}{7} (\bibinfo{year}{2016}), \bibinfo{pages}{1743--1756}.
\newblock


\bibitem[Morabit et~al\mbox{.}(2021)]%
        {morabit2021machine}
\bibfield{author}{\bibinfo{person}{Mouad Morabit}, \bibinfo{person}{Guy Desaulniers}, {and} \bibinfo{person}{Andrea Lodi}.} \bibinfo{year}{2021}\natexlab{}.
\newblock \showarticletitle{Machine-learning--based column selection for column generation}.
\newblock \bibinfo{journal}{\emph{Transportation Science}} \bibinfo{volume}{55}, \bibinfo{number}{4} (\bibinfo{year}{2021}), \bibinfo{pages}{815--831}.
\newblock


\bibitem[Muritiba et~al\mbox{.}(2010)]%
        {muritiba2010algorithms}
\bibfield{author}{\bibinfo{person}{Albert E~Fernandes Muritiba}, \bibinfo{person}{Manuel Iori}, \bibinfo{person}{Enrico Malaguti}, {and} \bibinfo{person}{Paolo Toth}.} \bibinfo{year}{2010}\natexlab{}.
\newblock \showarticletitle{Algorithms for the bin packing problem with conflicts}.
\newblock \bibinfo{journal}{\emph{Informs Journal on computing}} \bibinfo{volume}{22}, \bibinfo{number}{3} (\bibinfo{year}{2010}), \bibinfo{pages}{401--415}.
\newblock


\bibitem[Pferschy and Schauer(2009)]%
        {pferschy2009knapsack}
\bibfield{author}{\bibinfo{person}{Ulrich Pferschy} {and} \bibinfo{person}{Joachim Schauer}.} \bibinfo{year}{2009}\natexlab{}.
\newblock \showarticletitle{The knapsack problem with conflict graphs.}
\newblock \bibinfo{journal}{\emph{J. Graph Algorithms Appl.}} \bibinfo{volume}{13}, \bibinfo{number}{2} (\bibinfo{year}{2009}), \bibinfo{pages}{233--249}.
\newblock


\bibitem[Pferschy and Schauer(2017)]%
        {pferschy2017approximation}
\bibfield{author}{\bibinfo{person}{Ulrich Pferschy} {and} \bibinfo{person}{Joachim Schauer}.} \bibinfo{year}{2017}\natexlab{}.
\newblock \showarticletitle{Approximation of knapsack problems with conflict and forcing graphs}.
\newblock \bibinfo{journal}{\emph{Journal of Combinatorial Optimization}} \bibinfo{volume}{33}, \bibinfo{number}{4} (\bibinfo{year}{2017}), \bibinfo{pages}{1300--1323}.
\newblock


\bibitem[Platt(1999)]%
        {platt1999probabilistic}
\bibfield{author}{\bibinfo{person}{John Platt}.} \bibinfo{year}{1999}\natexlab{}.
\newblock \showarticletitle{Probabilistic outputs for support vector machines and comparisons to regularized likelihood methods}.
\newblock \bibinfo{journal}{\emph{Advances in large margin classifiers}} \bibinfo{volume}{10}, \bibinfo{number}{3} (\bibinfo{year}{1999}), \bibinfo{pages}{61--74}.
\newblock


\bibitem[Ram\'{\i}rez~S\'{a}nchez et~al\mbox{.}(2023)]%
        {QlearnACO}
\bibfield{author}{\bibinfo{person}{Jairo~Enrique Ram\'{\i}rez~S\'{a}nchez}, \bibinfo{person}{Camilo Chac\'{o}n~Sartori}, {and} \bibinfo{person}{Christian Blum}.} \bibinfo{year}{2023}\natexlab{}.
\newblock \showarticletitle{Q-Learning Ant Colony Optimization supported by Deep Learning for Target Set Selection}. In \bibinfo{booktitle}{\emph{Proceedings of the Genetic and Evolutionary Computation Conference}} (Lisbon, Portugal) \emph{(\bibinfo{series}{GECCO '23})}. \bibinfo{publisher}{Association for Computing Machinery}, \bibinfo{address}{New York, NY, USA}, \bibinfo{pages}{357–366}.
\newblock
\showISBNx{9798400701191}
\urldef\tempurl%
\url{https://doi.org/10.1145/3583131.3590396}
\showDOI{\tempurl}


\bibitem[Renault et~al\mbox{.}(2015)]%
        {renault2015online}
\bibfield{author}{\bibinfo{person}{Marc~P Renault}, \bibinfo{person}{Adi Ros{\'e}n}, {and} \bibinfo{person}{Rob van Stee}.} \bibinfo{year}{2015}\natexlab{}.
\newblock \showarticletitle{Online algorithms with advice for bin packing and scheduling problems}.
\newblock \bibinfo{journal}{\emph{Theoretical Computer Science}}  \bibinfo{volume}{600} (\bibinfo{year}{2015}), \bibinfo{pages}{155--170}.
\newblock


\bibitem[Ryan and Foster(1981)]%
        {ryan1981integer}
\bibfield{author}{\bibinfo{person}{David~M Ryan} {and} \bibinfo{person}{Brian~A Foster}.} \bibinfo{year}{1981}\natexlab{}.
\newblock \showarticletitle{An integer programming approach to scheduling}.
\newblock \bibinfo{journal}{\emph{Computer scheduling of public transport urban passenger vehicle and crew scheduling}} (\bibinfo{year}{1981}), \bibinfo{pages}{269--280}.
\newblock


\bibitem[Sadykov and Vanderbeck(2013)]%
        {BPPC}
\bibfield{author}{\bibinfo{person}{Ruslan Sadykov} {and} \bibinfo{person}{Fran\c{c}ois Vanderbeck}.} \bibinfo{year}{2013}\natexlab{}.
\newblock \showarticletitle{Bin Packing with Conflicts: A Generic Branch-and-Price Algorithm}.
\newblock \bibinfo{journal}{\emph{INFORMS Journal on Computing}} \bibinfo{volume}{25}, \bibinfo{number}{2} (\bibinfo{year}{2013}), \bibinfo{pages}{244--255}.
\newblock
\urldef\tempurl%
\url{https://doi.org/10.1287/ijoc.1120.0499}
\showDOI{\tempurl}
\showeprint{https://doi.org/10.1287/ijoc.1120.0499}


\bibitem[Schoenfield(2002)]%
        {Hard28}
\bibfield{author}{\bibinfo{person}{Jon~E Schoenfield}.} \bibinfo{year}{2002}\natexlab{}.
\newblock \showarticletitle{Fast, exact solution of open bin packing problems without linear programming}.
\newblock \bibinfo{journal}{\emph{Draft, US Army Space and Missile Defense Command, Huntsville, Alabama, USA}} (\bibinfo{year}{2002}).
\newblock


\bibitem[Shen et~al\mbox{.}(2023)]%
        {enhanceYZ}
\bibfield{author}{\bibinfo{person}{Yunzhuang Shen}, \bibinfo{person}{Yuan Sun}, \bibinfo{person}{Xiaodong Li}, \bibinfo{person}{Andrew Eberhard}, {and} \bibinfo{person}{Andreas Ernst}.} \bibinfo{year}{2023}\natexlab{}.
\newblock \showarticletitle{Enhancing column generation by a machine-learning-based pricing heuristic for graph coloring}.
\newblock \bibinfo{journal}{\emph{Proceedings of the AAAI Conference on Artificial Intelligence}}, \bibinfo{pages}{9}.
\newblock


\bibitem[Sun et~al\mbox{.}(2021)]%
        {usingstatistical}
\bibfield{author}{\bibinfo{person}{Yuan Sun}, \bibinfo{person}{Xiaodong Li}, {and} \bibinfo{person}{Andreas Ernst}.} \bibinfo{year}{2021}\natexlab{}.
\newblock \showarticletitle{Using Statistical Measures and Machine Learning for Graph Reduction to Solve Maximum Weight Clique Problems}.
\newblock \bibinfo{journal}{\emph{IEEE Transactions on Pattern Analysis and Machine Intelligence}} \bibinfo{volume}{43}, \bibinfo{number}{5} (\bibinfo{year}{2021}), \bibinfo{pages}{1746--1760}.
\newblock
\urldef\tempurl%
\url{https://doi.org/10.1109/TPAMI.2019.2954827}
\showDOI{\tempurl}


\bibitem[Sun et~al\mbox{.}(2022)]%
        {boosting}
\bibfield{author}{\bibinfo{person}{Yuan Sun}, \bibinfo{person}{Sheng Wang}, \bibinfo{person}{Yunzhuang Shen}, \bibinfo{person}{Xiaodong Li}, \bibinfo{person}{Andreas~T. Ernst}, {and} \bibinfo{person}{Michael Kirley}.} \bibinfo{year}{2022}\natexlab{}.
\newblock \showarticletitle{Boosting ant colony optimization via solution prediction and machine learning}.
\newblock \bibinfo{journal}{\emph{Computers \& Operations Research}}  \bibinfo{volume}{143} (\bibinfo{year}{2022}), \bibinfo{pages}{105769}.
\newblock
\showISSN{0305-0548}
\urldef\tempurl%
\url{https://doi.org/10.1016/j.cor.2022.105769}
\showDOI{\tempurl}


\bibitem[Talbi(2021)]%
        {talbi2021machine}
\bibfield{author}{\bibinfo{person}{El-Ghazali Talbi}.} \bibinfo{year}{2021}\natexlab{}.
\newblock \showarticletitle{Machine learning into metaheuristics: A survey and taxonomy}.
\newblock \bibinfo{journal}{\emph{ACM Computing Surveys (CSUR)}} \bibinfo{volume}{54}, \bibinfo{number}{6} (\bibinfo{year}{2021}), \bibinfo{pages}{1--32}.
\newblock


\bibitem[V{\'a}clav{\'i}k et~al\mbox{.}(2018)]%
        {vaclavikAcceleratingBranchandPriceAlgorithm2018}
\bibfield{author}{\bibinfo{person}{Roman V{\'a}clav{\'i}k}, \bibinfo{person}{Anton{\'i}n Nov{\'a}k}, \bibinfo{person}{P{\v r}emysl {\v S}{\r{u}}cha}, {and} \bibinfo{person}{Zden{\v e}k Hanz{\'a}lek}.} \bibinfo{year}{2018}\natexlab{}.
\newblock \showarticletitle{Accelerating the {{Branch-and-Price Algorithm Using Machine Learning}}}.
\newblock \bibinfo{journal}{\emph{European Journal of Operational Research}} \bibinfo{volume}{271}, \bibinfo{number}{3} (\bibinfo{date}{Dec.} \bibinfo{year}{2018}), \bibinfo{pages}{1055--1069}.
\newblock
\showISSN{03772217}
\urldef\tempurl%
\url{https://doi.org/10.1016/j.ejor.2018.05.046}
\showDOI{\tempurl}


\bibitem[Vanderbeck(2000)]%
        {vanderbeck2000}
\bibfield{author}{\bibinfo{person}{François Vanderbeck}.} \bibinfo{year}{2000}\natexlab{}.
\newblock \showarticletitle{{On Dantzig-Wolfe Decomposition in Integer Programming and ways to Perform Branching in a Branch-and-Price Algorithm}}.
\newblock \bibinfo{journal}{\emph{Operations Research}} \bibinfo{volume}{48}, \bibinfo{number}{1} (\bibinfo{date}{February} \bibinfo{year}{2000}), \bibinfo{pages}{111--128}.
\newblock
\urldef\tempurl%
\url{https://doi.org/10.1287/opre.48.1.111.124}
\showDOI{\tempurl}


\bibitem[Vanderbeck and Savelsbergh(2006)]%
        {vanderbeck2006generic}
\bibfield{author}{\bibinfo{person}{Fran{\c{c}}ois Vanderbeck} {and} \bibinfo{person}{Martin~WP Savelsbergh}.} \bibinfo{year}{2006}\natexlab{}.
\newblock \showarticletitle{A generic view of Dantzig--Wolfe decomposition in mixed integer programming}.
\newblock \bibinfo{journal}{\emph{Operations Research Letters}} \bibinfo{volume}{34}, \bibinfo{number}{3} (\bibinfo{year}{2006}), \bibinfo{pages}{296--306}.
\newblock


\bibitem[Xiang et~al\mbox{.}(2021)]%
        {xiang2021pairwise}
\bibfield{author}{\bibinfo{person}{Xiaoshu Xiang}, \bibinfo{person}{Ye Tian}, \bibinfo{person}{Xingyi Zhang}, \bibinfo{person}{Jianhua Xiao}, {and} \bibinfo{person}{Yaochu Jin}.} \bibinfo{year}{2021}\natexlab{}.
\newblock \showarticletitle{A pairwise proximity learning-based ant colony algorithm for dynamic vehicle routing problems}.
\newblock \bibinfo{journal}{\emph{IEEE transactions on intelligent transportation systems}} \bibinfo{volume}{23}, \bibinfo{number}{6} (\bibinfo{year}{2021}), \bibinfo{pages}{5275--5286}.
\newblock


\bibitem[Ye et~al\mbox{.}(2023)]%
        {ye2023deepaco}
\bibfield{author}{\bibinfo{person}{Haoran Ye}, \bibinfo{person}{Jiarui Wang}, \bibinfo{person}{Zhiguang Cao}, \bibinfo{person}{Helan Liang}, {and} \bibinfo{person}{Yong Li}.} \bibinfo{year}{2023}\natexlab{}.
\newblock \showarticletitle{DeepACO: Neural-enhanced Ant Systems for Combinatorial Optimization}. In \bibinfo{booktitle}{\emph{Advances in Neural Information Processing Systems}}.
\newblock


\bibitem[Zhang et~al\mbox{.}(2023)]%
        {zhang2023survey}
\bibfield{author}{\bibinfo{person}{Jiayi Zhang}, \bibinfo{person}{Chang Liu}, \bibinfo{person}{Xijun Li}, \bibinfo{person}{Hui-Ling Zhen}, \bibinfo{person}{Mingxuan Yuan}, \bibinfo{person}{Yawen Li}, {and} \bibinfo{person}{Junchi Yan}.} \bibinfo{year}{2023}\natexlab{}.
\newblock \showarticletitle{A survey for solving mixed integer programming via machine learning}.
\newblock \bibinfo{journal}{\emph{Neurocomputing}}  \bibinfo{volume}{519} (\bibinfo{year}{2023}), \bibinfo{pages}{205--217}.
\newblock


\end{thebibliography}

\end{document}